\documentclass{article}
\usepackage{iclr2027_conference,times}

\usepackage{amsmath,amsfonts,bm}

\def\eqref#1{equation~\ref{#1}}

\def\1{\bm{1}}

\DeclareMathAlphabet{\mathsfit}{\encodingdefault}{\sfdefault}{m}{sl}
\SetMathAlphabet{\mathsfit}{bold}{\encodingdefault}{\sfdefault}{bx}{n}

\usepackage{hyperref}
\usepackage{url}
\usepackage{graphicx}
\usepackage{booktabs}
\usepackage{longtable}
\usepackage{wrapfig}
\usepackage{placeins}
\usepackage{algorithm}
\usepackage{algpseudocode}
\title{Value-Guided Repulsion \\ for Offline Reinforcement Learning}
\author{%
  \textbf{Yixue Hao}, \ \textbf{Jiahao Zhang}, \ \textbf{Jiesong Lian}, \ \textbf{Xufei Guo}, \ \textbf{Xumeng Guo}, \\[2pt]
  \textbf{Wenhui Zhang}, \ \textbf{Long Hu}, \ \textbf{Rui Wang} \\[5pt]
  School of Computer Science and Technology, Huazhong University of Science and Technology \\[3pt]
  \texttt{\{yixuehao, ruiwang2020\}@hust.edu.cn}
}
\iclrfinalcopy 
\begin{document}
\maketitle
\begin{abstract}
Drifting models have emerged as a highly efficient and stable paradigm for 
policy design in offline reinforcement learning (RL).
However, standard drift policies determine repulsion solely from geometric
relationships among generated actions and therefore cannot distinguish them by
expected value. To address this limitation, we introduce Value-Guided Repulsion Q-Learning (VRQL) to integrate action value estimates into the particle 
interaction field. For each state, VRQL combines geometric kernel distance with 
a normalized Q value scaling. This mechanism assigns greater repulsive influence 
to particles with lower values while strictly retaining attraction toward behavior action. 
Consequently, this value-guided interaction field aligns behavior regularization
with policy improvement without adding an auxiliary behavior model stage. On 50 OGBench task 
configurations and 6 D4RL AntMaze configurations, VRQL demonstrates highly 
competitive performance against established baselines, suggesting that value-guided 
particle interaction is an effective method for offline RL.
\end{abstract}

\begin{wrapfigure}{R}{0.5\linewidth}
    \vspace{-25pt} %
    \centering
    \includegraphics[width=\linewidth]{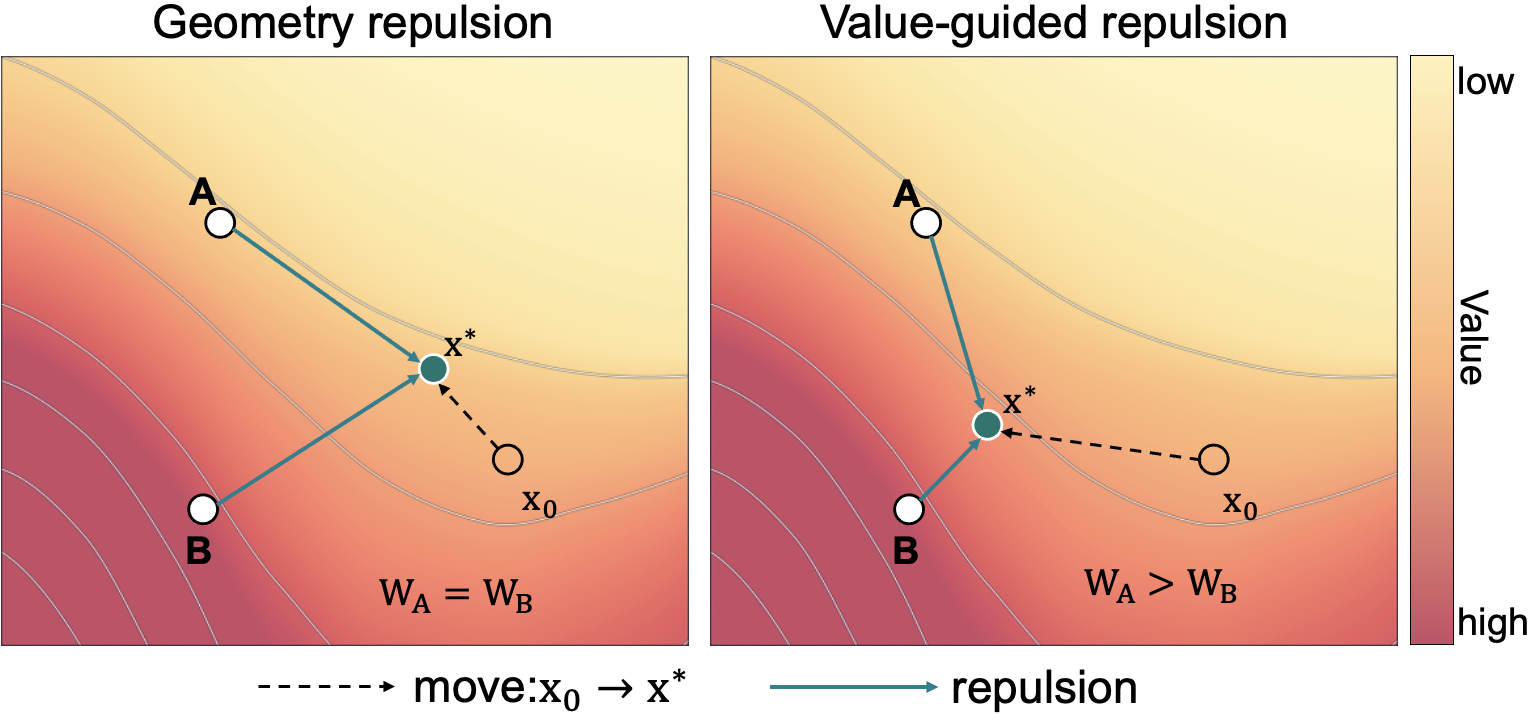}
    \caption{(Left) Standard
    drift policies rely on geometric distance, assigning identical
    repulsive weights ($W_A = W_B$) to equidistant neighbors particle A and B. (Right)
    VRQL incorporates critic estimates into the interaction field. The lower
    valued neighbor A exerts a stronger repulsive force ($W_A > W_B$), pushing
    the generated particle $x_0$ toward the higher value region at $x^*$.
    Background indicate the Q value landscape.}
    \label{fig:value-repulsion}
    \vspace{-10pt} %
\end{wrapfigure}

\section{Introduction}
\label{sec:introduction}

Offline reinforcement learning (RL) aims to learn effective policies from static 
datasets~\citep{levine2020offline}. As modern datasets scale and aggregate 
heterogeneous sources of varying quality, they exhibit complex and multimodal 
behavioral distributions. This complexity requires expressive policies to accurately model 
the data support and enforce behavioral constraints
~\citep{fujimoto2021minimalist,kostrikov2022offline,wang2023diffusion}.

Generative models, such as diffusion and flow matching approaches, provide the
expressiveness required to capture complex, multimodal action
distributions~\citep{ho2020denoising,lipman2023flow,liu2023flow,albergo2023building,tong2024improving} in offline RL. However, the iterative sampling process incurs high inference costs, 
and directly training flow or diffusion policies proves difficult due 
to unstable backpropagation through time~\citep{wang2023diffusion,park2025flow}. 
A common strategy adopts a two stage pipeline: first fitting a generative behavior model, and then using its samples to extract a separate actor~\citep{chen2023offline,hansenestruch2023idql,park2025flow}. 
Furthermore, because offline datasets contain suboptimal trajectories, 
these decoupled methods weight the training data using estimated Q values to bias the actor toward high return actions~\citep{peng2019advantage,ding2024diffusion,tiofack2026guided}. While effective at injecting value guidance, these two stage pipelines are inherently 
indirect and risk introducing compounding approximation errors.

By directly constructing the policy without relying on iterative sampling or 
auxiliary models, drifting models provide a promising alternative that sidesteps 
both inference inefficiencies and two stage approximation errors
~\citep{deng2026generative,houssaini2026drift}. These models train a conditional 
generator using an interaction field defined by attraction toward dataset actions 
and repulsion among generated particles to maintain diversity. 
However, while drifting models address the structural bottlenecks of two stage pipelines, they currently lack a mechanism for value-aware target shaping. Standard drift policies 
compute repulsive influence relying solely on geometric relationships. Consequently, 
particles at similar distances exert identical repulsive forces, even if their
corresponding critic estimates differ significantly
(Figure~\ref{fig:value-repulsion}, left). \textbf{Thus, a fundamental limitation of 
existing drift policies is their inability to incorporate estimated action 
values into the policy constraint, thereby hindering policy improvement.}

To address this limitation, we propose VRQL, which integrates
action values into the generated action interaction field. For each state, 
VRQL modulates the repulsion logits by combining kernel distance with a normalized 
value bias. Under this design, particles with lower values exert stronger repulsive 
influence, whereas those with higher values exert less forces
(Figure~\ref{fig:value-repulsion}, right), while the original attraction
toward behavior action remains strictly preserved. 
To ensure stable optimization, we adaptively calibrate this value signal using a shared scale derived from the 
typical within state value variation. Thus, VRQL maintains a direct value maximization 
objective while injecting value awareness into the particle interaction. This integration 
dynamically reshapes the behavioral constraint within the generative process, 
avoiding the need for an auxiliary behavior model.

In summary, our primary contributions are:
\begin{itemize}
\item \textbf{Value-Guided Repulsion:} We propose a value-guided 
repulsion mechanism. By leveraging estimated action values and geometric 
proximity, VRQL modulates the influence of generated neighbors particles within the drifting policy.
\item \textbf{Adaptive Value Calibration:} We introduce adaptive value calibration to control the strength of value guidance using the typical within state variation of particle values.
\item \textbf{Extensive Empirical Validation:} We evaluate VRQL across 50 OGBench tasks 
and 6 D4RL AntMaze environments~\citep{fu2020d4rl,park2025ogbench}, demonstrating 
that our method yields highly competitive performance in the offline RL setting.
\end{itemize}

\section{Preliminaries}
\label{sec:preliminaries}

\subsection{Policy Constrained Offline Reinforcement Learning}
\label{sec:offline-rl}

We consider a discounted Markov decision process
$(\mathcal S,\mathcal A,P,r,\gamma)$~\citep{sutton2018reinforcement}, where $\mathcal S$ and
$\mathcal A\subseteq\mathbb R^d$ are the state and action spaces, $P$ is the
transition kernel, $r$ is the reward function, and $\gamma\in[0,1)$ is the
discount factor. We assume $\mathcal A$ is a bounded action box and denote
coordinatewise projection onto it by $\Pi_{\mathcal A}$. Given an
initial state distribution, a policy $\pi$ seeks to maximize
$J(\pi)=\mathbb E_\pi[\sum_{t=0}^{\infty}\gamma^t r(s_t,a_t)]$.
Its action value function $Q^\pi(s,a)$ is the expected discounted return
conditioned on starting with $(s_0,a_0)=(s,a)$ and subsequently following $\pi$.

Offline learning uses only a fixed dataset
$\mathcal D=\{(s,a,r,s',m)\}$ collected by a behavior policy $\mu$, where
$m$ is a continuation mask~\citep{levine2020offline}. A policy constrained
offline objective combines value improvement with a soft behavioral regularizer:
\begin{equation}
 \max_{\pi}\quad
 \mathbb E_{s\sim\mathcal D,\,a\sim\pi(\cdot\mid s)}
 \left[Q^\pi(s,a)\right]
 -\lambda\,\mathcal R_{\mathcal D}(\pi,\mu),
 \qquad \lambda>0.
 \label{eq:pcorl-objective}
\end{equation}
Here $\mathcal R_{\mathcal D}$ denotes a behavioral regularizer supported
by the dataset; it need not be a hard equality or a fixed divergence to $\mu$.
This soft formulation permits controlled movement within the learned action
distribution while discouraging unsupported actions whose critic estimates may
be unreliable~\citep{fujimoto2019off,kumar2020conservative}.

In practice, the value term is estimated with an ensemble of $H$ critics, with
$[H]=\{1,\ldots,H\}$, online parameters $\{\phi_h\}_{h\in[H]}$, and target
parameters $\{\bar\phi_h\}_{h\in[H]}$. For a next-state action sampled from the current actor,
$x_\theta(s',z')$, we define the target as
\begin{equation}
 y=r+\gamma m\min_{h\in[H]}
 Q_{\bar\phi_h}(s',x_\theta(s',z')).
 \label{eq:critic-target}
\end{equation}
The corresponding Bellman regression is
\begin{equation}
 \mathcal L_{\mathrm{critic}}
 =\frac{1}{H}\sum_{h\in[H]}
 \mathbb E_{(s,a,r,s',m)\sim\mathcal D,\,z'}
 \left[(Q_{\phi_h}(s,a)-y)^2\right].
 \label{eq:critic-objective}
\end{equation}
For actor updates, let
$\mathcal Q_\phi(s,a)=\operatorname{Agg}_{h\in[H]}Q_{\phi_h}(s,a)$ denote the
chosen ensemble aggregation. The target critics are updated
by Polyak averaging. VRQL leaves these standard value estimation components
unchanged and modifies the behavioral regularizer $\mathcal R_{\mathcal D}$.

\subsection{Drifting Models}
\label{sec:drifting-background}

Drifting models train a direct conditional generator
$x_\theta(s,z)=\Pi_{\mathcal A}(f_\theta(s,z))$, where
$z\sim\mathcal N(0,I)$, by regressing its outputs toward stopped targets
constructed from attractive and repulsive sample interactions~\citep{deng2026generative}. The noise induces the stochastic policy
$\pi_\theta(\cdot\mid s)$, although the generator is deterministic for fixed
$(s,z)$. Given generated particles $x_i$ and a set of positive samples
$\mathcal P(s)$, a positive kernel $k_\tau$ with bandwidth parameter $\tau$ produces normalized attraction and
repulsion weights. A common choice is a Laplace or Gaussian kernel scaled by the action dimension $d$,
\begin{equation}
 k_\tau^{\mathrm L}(x,y)=\exp\!\left(-\frac{\|x-y\|_2}{\tau\sqrt d}\right),
 \qquad
 k_\tau^{\mathrm G}(x,y)=\exp\!\left(-\frac{\|x-y\|_2^2}{2\tau^2 d}\right).
 \label{eq:kernels}
\end{equation}
With $w^+_{iu} \propto k_\tau(x_i, u)$ normalized over $u\in\mathcal P(s)$ and $w^-_{ij} \propto k_\tau(x_i, x_j)$ normalized
over $j\ne i$, the total drift vector $V_i$ aggregates the attraction vector $A_i$ pulling toward the behavior data, and the repulsion vector $R_i$ pushing away from neighboring particles:
\begin{equation}
 A_i=\sum_{u\in\mathcal P(s)}w^+_{iu}(u-x_i),\qquad
 R_i=\sum_{j\ne i}w^-_{ij}(x_j-x_i),\qquad
 V_i=A_i-R_i.
 \label{eq:generic-drift}
\end{equation}
The generator is trained toward a projected, stopped target $t_i$:
\begin{equation}
 t_i=\operatorname{sg}\!\left[\Pi_{\mathcal A}(x_i+\eta V_i)\right],
 \qquad
 \mathcal L_{\mathrm{drift}}(\theta)
 =\mathbb E\left[\frac{1}{d}\|x_i-t_i\|_2^2\right],
 \label{eq:standard-drift-objective}
\end{equation}
where $\eta>0$ is a drift step size. The targets are recomputed as training
proceeds and held fixed when differentiating the regression loss. Standard
drifting models use geometry alone in $w^-_{ij}$; VRQL changes this negative side
weighting while retaining the attraction construction.

\section{Methods}
\label{sec:methods}

Our method VRQL builds on drifting policies: it retains the actor network and behavior 
action anchor while redefining how generated particles repel each other. Algorithm~\ref{alg:vrql} summarizes the complete training procedure.
Specifically, for a given state, the actor generates a particle set, and the critic evaluates these particles. Then, a value signal combines with pairwise geometric distances to determine the repulsion weights. The resulting interaction field forms the drift target for actor optimization. Thus, VRQL reshapes the behavioral
target through value-guided interactions while preserving the value
maximization objective~\citep{houssaini2026drift}.

In this section, we consider a drift minibatch containing $B_d$ transitions and use
$b\in\{1,\cdots,B_d\}$ to index the transitions .

\subsection{Behavior Action Anchor and Particle Generation}
\label{sec:policy-anchor}

For each state $s_b$, we sample $N$ independent noise vectors $z_b^i$ and
generate the particle set $\{x_{b}^{i}\}$:
\begin{equation}
 x_b^i = x_\theta(s_b,z_b^i),
 \qquad
 i\in\{1,\cdots,N\}.
 \label{eq:particle-generation}
\end{equation}
The dataset behavior action provides a positive anchor 
$a_b^+ = \Pi_{\mathcal A}(a_b)$. The positive set in Eq.~\ref{eq:generic-drift} 
reduces to the singleton $\mathcal P(s_b) = \{a_b^+\}$, yielding the normalized 
attraction field $A_{b}^{i}$:
\begin{equation}
 A_b^i = a_b^+ - x_b^i.
 \label{eq:singleton-attraction}
\end{equation}
At inference, the actor outputs actions directly. Value guidance acts only 
on generated neighbor interactions during training, leaving the attraction 
mechanism and inference architecture unchanged.

\subsection{Adaptive Value Calibration}
\label{sec:adaptive-calibration}

We define the value used to weight particle $j$ as:
\begin{equation}
 q_b^j = \operatorname{sg}\!\left[\operatorname{Agg}_{h\in[H]}
 Q_{\bar\phi_h}(s_b,x_b^j)\right].
 \label{eq:particle-values}
\end{equation}
VRQL uses the target critic ensemble with minimum aggregation to compute this 
signal. The critic aggregation rules are configurable, as specified 
in Appendix~\ref{app:experimental-details}.

Because raw Q value magnitudes vary significantly across states and environments,
using uncalibrated values makes the relative strength of value guidance sensitive. 
To ensure stable optimization, we center the Q values within each state 
using the mean $\bar q_b = \frac{1}{N}\sum_{j=1}^{N}q_b^j$ and apply a shared 
scale for the centered differences across the minibatch. The default scale,
$S_{\mathrm{norm}}=S_{\mathrm{std}}$, is the average within state standard deviation:
\begin{equation}
 S_{\mathrm{std}} =
 \frac{1}{B_d}\sum_{b=1}^{B_d}\sqrt{\frac{1}{N}\sum_{j=1}^{N}
 (q_b^j-\bar q_b)^2} + \epsilon ,
 \label{eq:value-scale}
\end{equation}
where $\epsilon$ is a constant added for numerical stability.
Centering subtracts the same value from all particle logits within each
state and therefore leaves the softmax weights unchanged. We use it to
remove common value offsets before computing the logits. The shared
scale $S_{\mathrm{std}}$ calibrates the strength of value guidance relative
to typical within state value variation. Recomputing this scale at each
training step adapts the calibration as the critic evolves. A shared
scale also preserves differences in value dispersion across states,
whereas normalizing each state separately can amplify small value
differences in states with low variance.
Section~\ref{sec:normalization-ablation} evaluates alternative scaling
schemes.

\subsection{Value-Guided Repulsion}
\label{sec:value-repulsion}

We measure geometric proximity using the Laplace kernel in Eq.~\ref{eq:kernels} by default, denoted as $k_\tau$ throughout.
This geometric term is combined with the calibrated value signal to define
the repulsion logit for each distinct pair $j \ne i$:
\begin{equation}
 \ell_b^{i,j} =
 \underbrace{-\frac{\lVert x_b^i-x_b^j\rVert_2}{\tau\sqrt d}}_{\text{geometry}}
 - \underbrace{\frac{q_b^j-\bar q_b}{\beta S_{\mathrm{norm}}}}_{\text{value}},
 \qquad
 w_b^{i,j} = \frac{\exp(\ell_b^{i,j})}
 {\sum_{j'\ne i}\exp(\ell_b^{i,j'})},
 \quad w_b^{i,i} = 0,
 \label{eq:repulsion-weights}
\end{equation}
where $\beta > 0$ is the guidance temperature. With this scaling,
$\beta$ controls the strength of value guidance relative to typical within state
value variation. By substituting the standard geometric weights $w^-_{ij}$ in 
Eq.~\ref{eq:generic-drift} with these value-guided weights $w_b^{i,j}$, we define 
the value-guided repulsive component $R_b^i$ and the combined drift vector $V_b^i$ as:
\begin{equation}
 R_b^i = \sum_{j\ne i}w_b^{i,j}(x_b^j-x_b^i),
 \qquad V_b^i = A_b^i - R_b^i.
 \label{eq:value-drift}
\end{equation}
The diagonal mask prevents self interaction. Since the normalized weights 
sum to one, the combined drift simplifies to 
$V_b^i = a_b^+ - \sum_{j\ne i}w_b^{i,j}x_b^j$. Value guidance shifts the 
neighbor barycenter that determines the update direction. 
Although the repulsion weights sum to one, value-guided 
reweighting can change both the direction and magnitude of the resulting drift.
Specifically, consider two eligible neighbors $(j, k \ne i)$, their relative repulsive influence 
on particle $i$ is:
\begin{equation}
 \frac{w_b^{i,j}}{w_b^{i,k}} =
 \frac{k_\tau(x_b^i,x_b^j)}{k_\tau(x_b^i,x_b^k)}
 \exp\!\left(-\frac{q_b^j-q_b^k}{\beta S_{\mathrm{norm}}}\right).
 \label{eq:relative-influence}
\end{equation}
This formulation combines geometric proximity with value differences. At equal distances, the 
lower valued neighbor receives greater repulsive weight. At unequal distances, 
spatial proximity and critic evaluations determine the influence. When particle 
values are identical or as $\beta \to \infty$, the method recovers geometric 
repulsion.
\paragraph{Actor optimization objective.}
The projected drift target $t_b^i = \Pi_{\mathcal A}(x_b^i+\eta V_b^i)$ is
stopped as in Eq.~\ref{eq:standard-drift-objective}. The actor loss is:
\begin{equation}
 \mathcal L_{\mathrm{actor}}(\theta)
 = \alpha\,\mathcal L_{\mathrm{drift}}(\theta)
 - \mathbb E_{s\sim\mathcal D,z\sim\mathcal N(0,I)}
 \left[\mathcal Q_\phi(s,x_\theta(s,z))\right],
 \label{eq:actor-objective}
\end{equation}
where $\alpha > 0$ balances behavioral regularization and value improvement. 
The Q maximization term computes gradients through the online critic ensemble, 
whereas the values in Eq.~\ref{eq:particle-values} are detached before entering 
the repulsion weights. This stop gradient operation is necessary to prevent 
gradient paths from flowing through the drift interaction field or its value bias. 
The direct value objective improves actions through critic gradients, whereas
value-guided repulsion uses relative critic estimates to reshape the stopped
behavioral target. The two terms therefore incorporate value information through
different update mechanisms.

\paragraph{Computational overhead.}
Value guidance adds one additional batch of critic evaluations for the $B_dN$ generated 
actions to the $\mathcal O(B_dN^2d)$ pairwise interactions.

\begin{algorithm}[t]
\caption{Value-Guided Repulsion Q-Learning (VRQL)}
\label{alg:vrql}
\begin{algorithmic}[1]
\item[\textbf{Input:}] Dataset $\mathcal D$; policy $x_\theta$; critic ensemble $\{Q_{\phi_h}\}_{h\in[H]}$; Laplace kernel temperature $\tau$;
 guidance temperature $\beta$; particle count $N\ge 3$
\For{each training step}
 \State Sample a minibatch from $\mathcal D$ and update the critics $\{\phi_h\}_{h\in[H]}$ 
 \State Generate particles $x_b^i\leftarrow x_\theta(s_b,z_b^i)$, \ $z_b^i\sim\mathcal N(0,I)$, \ $i\in\{1,\cdots,N\}$
 \State Compute $q_b^j$, $\bar q_b$, and
    $S_{\mathrm{norm}}$ as in
    Section~\ref{sec:adaptive-calibration}
 \State $\ell_b^{i,j}\leftarrow-\lVert x_b^i-x_b^j\rVert_2/(\tau\sqrt d)-(q_b^j-\bar q_b)/(\beta S_{\mathrm{norm}})$, \ $j\ne i$ \Comment{value guidance}
 \State $w_b^{i,j}\leftarrow\operatorname{softmax}_{j\ne i}(\ell_b^{i,j})$, \ $w_b^{i,i}\leftarrow 0$
 \State $V_b^i\leftarrow(a_b^+-x_b^i)-\sum\nolimits_{j\ne i}w_b^{i,j}(x_b^j-x_b^i)$ \Comment{attraction $-$ repulsion}
 \State $t_b^i\leftarrow\operatorname{sg}\!\left[\Pi_{\mathcal A}(x_b^i+\eta V_b^i)\right]$ 
 \State $\mathcal L_{\mathrm{drift}}\leftarrow
\mathbb E_{b,i}\!\left[
\frac{1}{d}\lVert x_b^i-t_b^i\rVert_2^2
\right]$
 \State Update $\theta$ by minimizing $\alpha\,\mathcal L_{\mathrm{drift}}-\mathbb E[\mathcal Q_\phi(s,x_\theta(s,z))]$ 
 \State Update the target critics $\{\bar\phi_h\}_{h\in[H]}$ by Polyak averaging
\EndFor
\item[\textbf{Output:}] One step policy $\pi_\theta$ with action $a=x_\theta(s,z)$
\end{algorithmic}
\end{algorithm}

\section{Experiments}
\label{sec:experiments}

In this section, we evaluate the policy performance against published baselines and investigate how
repulsion assignment,  normalization, guidance
temperature, and particle count affect VRQL.

\subsection{Experimental Setup}
\label{sec:experimental-setup}

\paragraph{Benchmarks.}
We evaluate VRQL on OGBench and D4RL AntMaze~\citep{fu2020d4rl,park2025ogbench}. 
OGBench offers 50 single task configurations across 10 categories, each comprising 5 tasks. These 
environments cover diverse domains spanning locomotion and navigation 
(\textit{humanoidmaze-medium/large-navigate}, \textit{antmaze-large/giant-navigate}, 
\textit{antsoccer-arena-navigate}), high dimensional object manipulation 
(\textit{cube-single-play}, \textit{cube-double-play}, and \textit{scene-play}), 
and long horizon puzzle solving (\textit{puzzle-3x3-play} and \textit{puzzle-4x4-play}). 
D4RL AntMaze adds 6 configurations covering three maze 
sizes (\textit{umaze}, \textit{medium}, \textit{large}) and two offline dataset variants (\textit{play}, \textit{diverse}), 
explicitly testing the ability to stitch suboptimal trajectories. Detailed hyperparameter 
configurations are provided in Appendix~\ref{app:hyperparameters}.

\paragraph{Baselines.}
We compare VRQL against five classes of offline algorithms: (i) behavior cloning and 
Gaussian policies, including BC, IQL~\citep{kostrikov2022offline}, and ReBRAC
~\citep{tarasov2023revisiting}; (ii) diffusion and consistency models, represented 
by IDQL~\citep{hansenestruch2023idql}, SRPO, and CAC; (iii) standard flow policies, 
comprising FQL~\citep{park2025flow}, FAWAC, FBRAC, and IFQL; (iv) value-guided policies, 
such as Q-Flow~\citep{doo2026qflow} and GFP~\citep{tiofack2026guided}, to benchmark 
alternative value integration strategies; and (v) the direct drift predecessor, 
DriftQL~\citep{houssaini2026drift}, which relies strictly on geometric repulsion to 
isolate the empirical contribution of our value-guided mechanism.

Owing to space constraints, Table~\ref{tab:main_results} reports only one
representative method from each baseline class: IQL,
IDQL, FQL, Q-Flow, and DriftQL. Appendix~\ref{app:benchmark-results} presents the per task scores for these together with fourteen methods (i.e., BC, ReBRAC, SRPO, CAC, FAWAC, FBRAC, IFQL and GFP). Baseline scores follow the corresponding papers~\citep{houssaini2026drift, doo2026qflow, tiofack2026guided}.

\paragraph{Evaluation.}
For all VRQL evaluations, we train with five independent random seeds per task
and run 50 evaluation episodes per checkpoint. OGBench scores correspond to average success rates computed at 800k, 900k, and 1M training steps within each seed; D4RL AntMaze scores use normalized returns measured at 500k training steps. We report per task means and standard deviations across the five
seed level scores. Table~\ref{tab:main_results} equally weights all tasks per OGBench category 
and D4RL AntMaze configuration. Category standard deviations aggregate task-level seed variances. 
For D4RL IQL and IDQL, only means are reported, as original papers do not standard deviations.

\subsection{Main Results}
\label{sec:main-results}

Table~\ref{tab:main_results} evaluates VRQL on OGBench and D4RL, 
demonstrating its empirical advantages across three distinct aspects. 
First, compared with purely geometric predecessor DriftQL, VRQL yields 
substantial gains, raising success rates from 60 to 74 on antmaze-giant 
and from 25 to 58 on cube-double. Moreover,  VRQL achieves
higher mean scores than IQL, IDQL and FQL across all evaluated task categories. 
These results indicate that value guidance improving performance in these environments. 
Then, VRQL outperforms the alternative value-guided
method Q-Flow on most task categories. These comparisons suggest that modulating the particle interaction
field is a promising strategy for incorporating value guidance into policy optimization. 
Finally, on the established D4RL AntMaze benchmark, VRQL reaches a mean 
score of 90, exceeding all evaluated  policy 
classes, which verifies its efficacy across classical offline continuous 
control domains.

\begin{table}[t]
\centering
\caption{\textbf{Offline reinforcement learning results.} VRQL achieves the best or tied for best performance across the OGBench categories and the D4RL AntMaze benchmark. The reported $\pm$ values aggregate the seed standard deviations across individual tasks. This table presents one representative method from each policy class, while Appendix~\ref{app:benchmark-results} provides complete results for all fourteen methods. Boldface indicates the best score in each row.}
\label{tab:main_results}

\vspace{\baselineskip}
\setlength{\tabcolsep}{3pt}
\begin{tabular}{@{} l *{6}{c} @{}}
\toprule
& \multicolumn{1}{c}{Gaussian} & \multicolumn{1}{c}{Diffusion} & \multicolumn{1}{c}{Flow} & \multicolumn{1}{c}{Value-Guide} & \multicolumn{2}{c}{Drift} \\
\cmidrule(lr){2-2} \cmidrule(lr){3-3} \cmidrule(lr){4-4} \cmidrule(lr){5-5} \cmidrule(lr){6-7}
Task category & IQL & IDQL & FQL & Q-Flow & DriftQL & \textbf{VRQL} \\
\midrule
\multicolumn{7}{@{}l}{\textit{OGBench}} \\
antmaze-large (5 tasks) & $53 \pm 3$ & $21 \pm 5$ & $79 \pm 3$ & $88 \pm 5$ & $92 \pm 4$ & $\textbf{97} \pm 1$ \\
antmaze-giant (5 tasks) & $4 \pm 1$ & $0 \pm 0$ & $9 \pm 6$ & $40 \pm 6$ & $60 \pm 2$ & $\textbf{74} \pm 4$ \\
humanoidmaze-medium (5 tasks) & $33 \pm 2$ & $1 \pm 0$ & $58 \pm 5$ & $83 \pm 4$ & $62 \pm 2$ & $\textbf{88} \pm 8$ \\
humanoidmaze-large (5 tasks) & $2 \pm 1$ & $1 \pm 0$ & $4 \pm 2$ & $\textbf{8} \pm 2$ & $5 \pm 8$ & $\textbf{8} \pm 1$ \\
antsoccer-arena (5 tasks) & $8 \pm 2$ & $12 \pm 4$ & $60 \pm 2$ & $55 \pm 5$ & $65 \pm 2$ & $\textbf{71} \pm 2$ \\
cube-single (5 tasks) & $83 \pm 3$ & $95 \pm 2$ & $96 \pm 1$ & $95 \pm 2$ & $94 \pm 3$ & $\textbf{99} \pm 1$ \\
cube-double (5 tasks) & $6 \pm 1$ & $15 \pm 6$ & $29 \pm 2$ & $36 \pm 5$ & $25 \pm 2$ & $\textbf{58} \pm 3$ \\
scene (5 tasks) & $28 \pm 1$ & $46 \pm 3$ & $56 \pm 2$ & $60 \pm 2$ & $74 \pm 4$ & $\textbf{75} \pm 1$ \\
puzzle-3x3 (5 tasks) & $9 \pm 1$ & $10 \pm 2$ & $30 \pm 1$ & $49 \pm 5$ & $35 \pm 3$ & $\textbf{54} \pm 2$ \\
puzzle-4x4 (5 tasks) & $7 \pm 1$ & $29 \pm 3$ & $17 \pm 2$ & $30 \pm 3$ & $27 \pm 3$ & $\textbf{32} \pm 2$ \\
\midrule
\multicolumn{7}{@{}l}{\textit{D4RL}} \\
antmaze (6 tasks) & $57$ & $79$ & $84 \pm 3$ & $59 \pm 1$ & $84 \pm 3$ & $\textbf{90} \pm 3$ \\
\bottomrule
\end{tabular}
\end{table}

\subsection{Ablation Studies}
\label{sec:ablations}

We evaluate four ablation studies, i.e., repulsion weight assignment, value normalization, guidance temperature, and particle count. 
To validate that VRQL generalizes across diverse environments, 
we distribute these evaluations across various tasks from both OGBench and D4RL, 
following the evaluation protocol of Section~\ref{sec:experimental-setup}.
The first three use OGBench tasks, the repulsion weight assignment and guidance temperature
studies each cover three tasks. While the particle count study adopts three D4RL AntMaze
configurations and additionally reports relative training time. All
configuration is trained with five independent seeds, and all figures show seed
means with shading denoting one standard deviation. Each ablation study
varies a single quantity and inherits all remaining hyperparameters from the
corresponding task configuration.

\begin{figure}[t]
 \centering
 \includegraphics[width=0.6\linewidth]{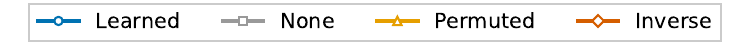}\par
 \begin{minipage}{0.33\linewidth}
 \centering
 \includegraphics[width=\linewidth]{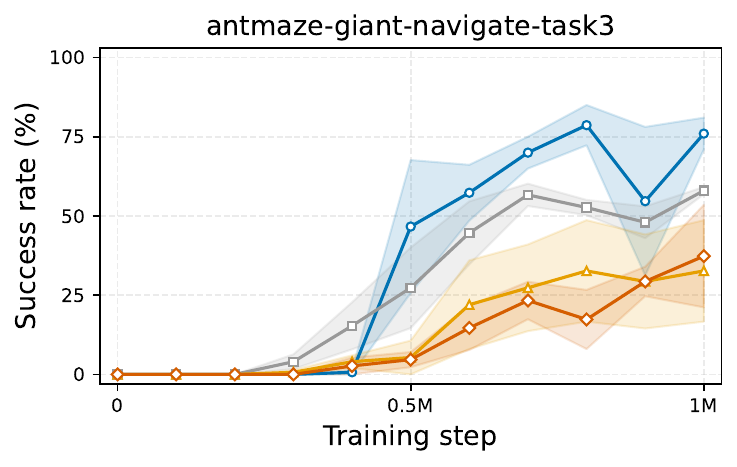}
 \end{minipage}\hfill
 \begin{minipage}{0.33\linewidth}
 \centering
 \includegraphics[width=\linewidth]{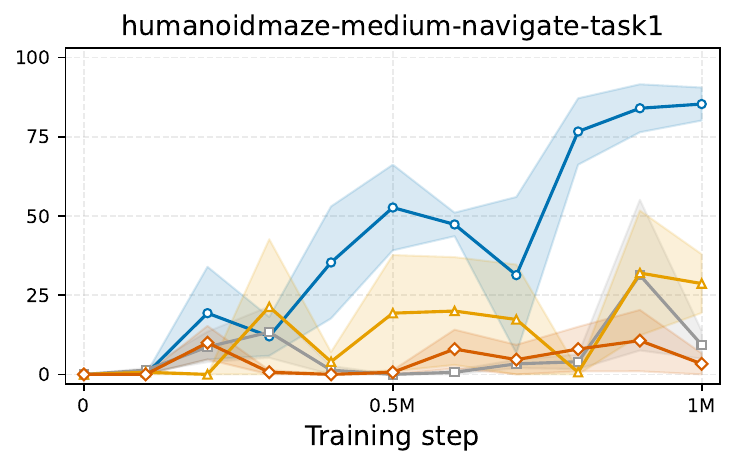}
 \end{minipage}\hfill
 \begin{minipage}{0.33\linewidth}
 \centering
 \includegraphics[width=\linewidth]{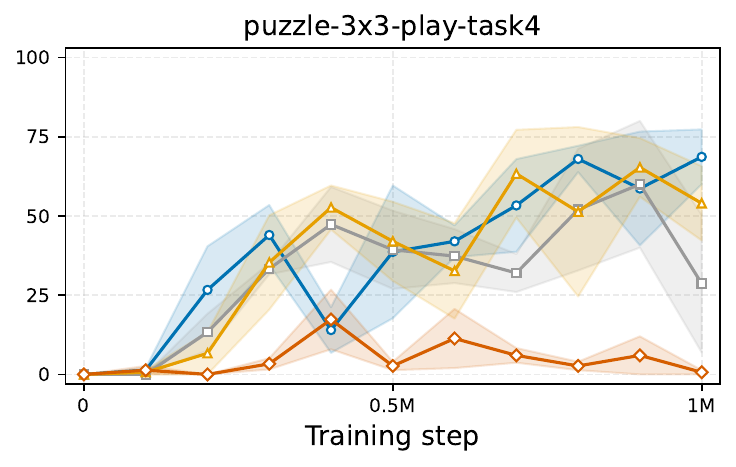}
 \end{minipage}\hfill
 \caption{\textbf{Effect of repulsion weight assignment.} The
\textit{learned} assignment used by VRQL is compared with geometric repulsion
(\textit{none}), shuffled value biases (\textit{permuted}), and reversed
value biases (\textit{inverse}). Curves show mean success rates, 
 and shaded regions denote one standard deviation. }
 \label{fig:assignment-ablation}
\end{figure}

\paragraph{Repulsion weight assignment.}
Figure~\ref{fig:assignment-ablation} evaluates the \textit{learned} assignment used by VRQL on three
single tasks (\textit{antmaze-giant-navigate}, \textit{humanoidmaze-medium-navigate}, \textit{puzzle-3x3-play}). We compare this mechanism against three control variants: \textit{none} (pure geometric repulsion), 
\textit{permuted} (shuffled value biases among particles within each state before the softmax operation), and \textit{inverse} (reversed value
biases). 

This ablation investigates whether the correspondence between estimated 
action quality and repulsive influence drives policy performance. 
The Figure~\ref{fig:assignment-ablation} shows that the \textit{learned} guidance achieves the highest success rates and fastest convergence across all three tasks. 
Moreover, disrupting the correct value relationship actively degrades learning, and 
the \textit{inverse} assignment cause severe performance collapse, plunging 
success rates near zero in humanoid navigation and puzzle solving. Meanwhile, 
the \textit{permuted} assignment  performs worse than the \textit{none} 
baseline, demonstrating that random value noise actively disrupts the beneficial 
geometric structure. These outcomes confirm that the precise alignment of high 
repulsion with low value actions is essential.

\FloatBarrier
\begin{wrapfigure}{r}{0.42\linewidth}
    \vspace{-20pt} %
    \centering

    \includegraphics[width=0.92142857\linewidth]{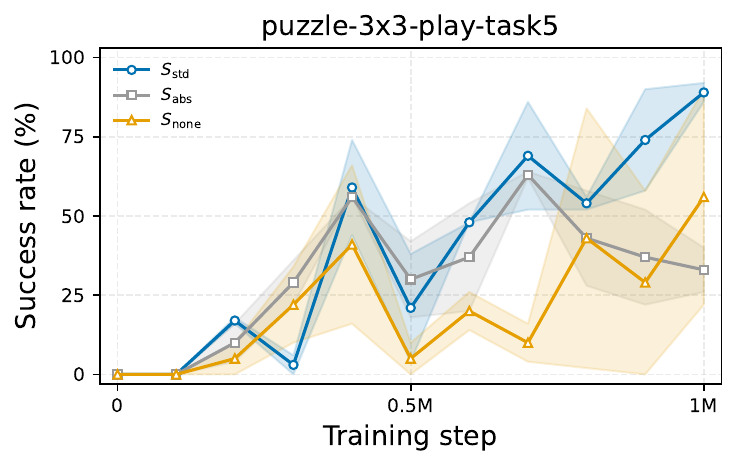}
    \caption{\textbf{Value normalization ablation.} Success rates under different normalization strategies. }
    \label{fig:normalization-ablation}
    \vspace{-18pt} %
\end{wrapfigure}

\paragraph{Value normalization.}
\label{sec:normalization-ablation}
Figure~\ref{fig:normalization-ablation} compares three scaling 
strategies for $S_{\mathrm{norm}}$: the default within state 
standard deviation $S_{\mathrm{std}}$ defined in 
Eq.~\ref{eq:value-scale}, the absolute mean $S_{\mathrm{abs}}$ define in Eq.~\ref{eq:normalization-ablation}, 
and the unnormalized baseline $S_{\mathrm{none}}$ (i.e., $S_{\mathrm{none}}=1$). 
\begin{equation}
 S_{\mathrm{abs}} =
 \frac{1}{B_dN}\sum_{b=1}^{B_d}\sum_{j=1}^{N}|q_b^j|+\epsilon.
 \label{eq:normalization-ablation}
\end{equation}
The results show that calibration by relative value variation using
$S_{\mathrm{std}}$ yields the highest and most stable 
success rate. Without normalization $S_{\mathrm{none}}$, 
training exhibits severe instability and variance, 
demonstrating that raw value scales disrupt the guidance 
signal. While absolute value scaling $S_{\mathrm{abs}}$ 
initially matches $S_{\mathrm{std}}$, it suffers from  
performance degradation later in training state. Thus, calibrating the value term with the within state value spread provides a more reliable guidance scale.

\begin{figure}[h]
 \centering
 \includegraphics[width=0.6\linewidth]{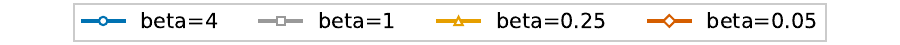}\par
 \begin{minipage}{0.33\linewidth}
 \centering
 \includegraphics[width=\linewidth]{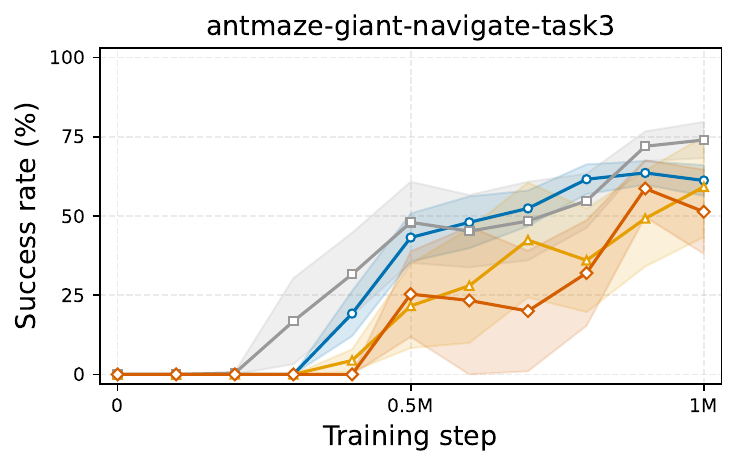}
 \end{minipage}\hfill
 \begin{minipage}{0.33\linewidth}
 \centering
 \includegraphics[width=\linewidth]{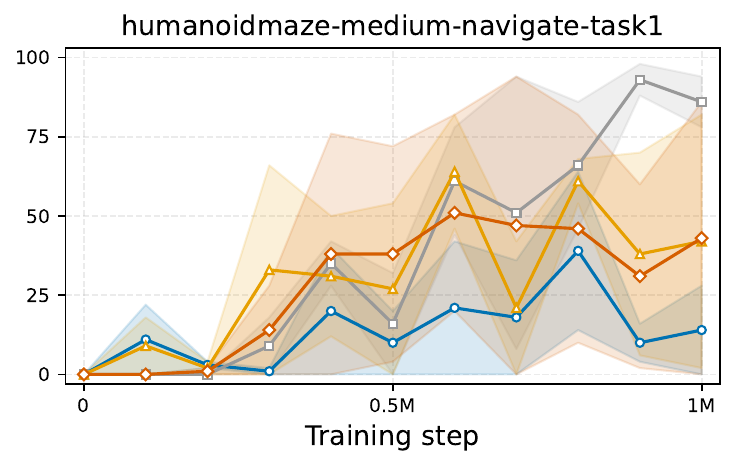}
 \end{minipage}\hfill
 \begin{minipage}{0.33\linewidth}
 \centering
 \includegraphics[width=\linewidth]{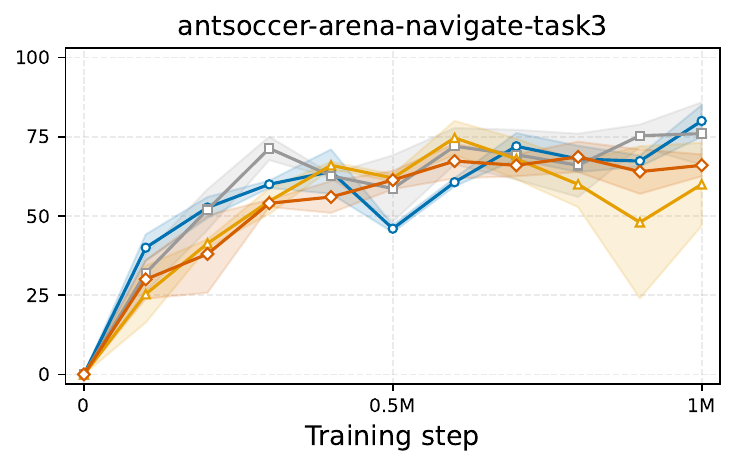}
 \end{minipage}\hfill
 \caption{\textbf{Sensitivity to the guidance temperature $\beta$.} Training curves for 
 varying $\beta$ on three OGBench tasks. Curves show mean success rates, 
 and shaded regions denote one standard deviation.}
 \label{fig:guidance-sensitivity}
\end{figure}

\paragraph{Guidance temperature.}
\label{sec:guidance-ablation}
Figure~\ref{fig:guidance-sensitivity} evaluates VRQL sensitivity to the guidance 
temperature $\beta \in \{0.05, 0.25, 1, 4\}$  on three single tasks (\textit{antmaze-giant-navigate}, \textit{humanoidmaze-medium-navigate}, 
\textit{antsoccer-arena-navigate}). Results indicate that 
low temperatures ($\beta=0.05$) disrupt geometric diversity and degrade success rates in 
antmaze-giant. Conversely, high temperatures ($\beta=4$) weaken the value signal and 
reduce performance in humanoid navigation. Thus, a moderate $\beta$ balances 
geometric diversity with value guidance.

\begin{figure}[t]
 \centering
  \includegraphics[width=0.6\linewidth]{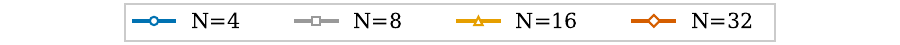}\par
 \begin{minipage}{0.33\linewidth}
 \centering
 \includegraphics[width=\linewidth]{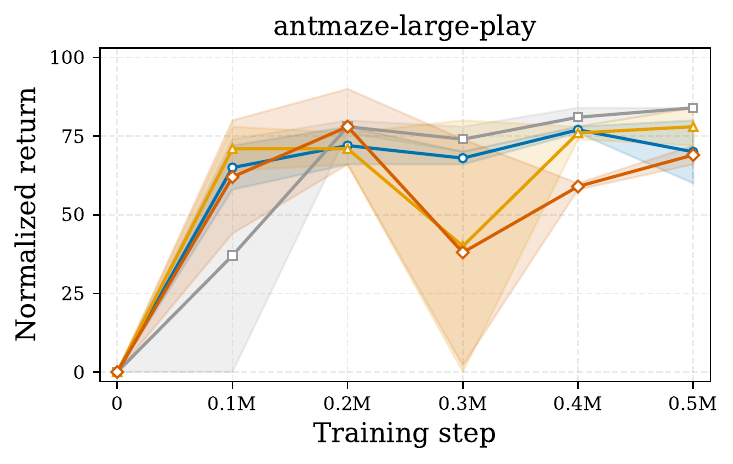}
 \end{minipage}\hfill
 \begin{minipage}{0.33\linewidth}
 \centering
 \includegraphics[width=\linewidth]{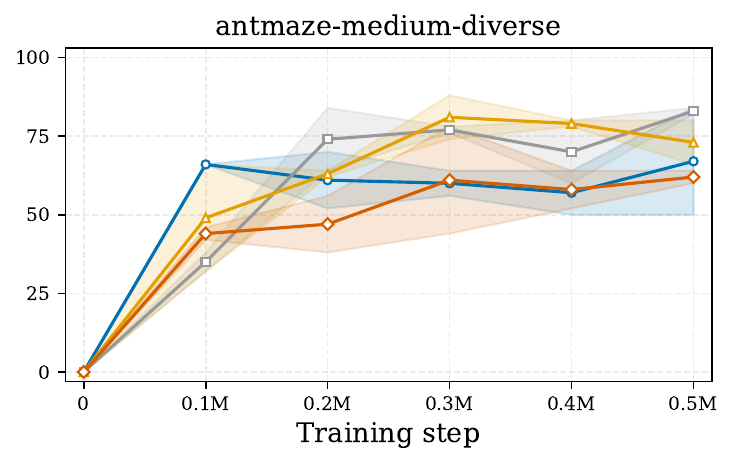}
 \end{minipage}\hfill
 \begin{minipage}{0.33\linewidth}
 \centering
 \includegraphics[width=\linewidth]{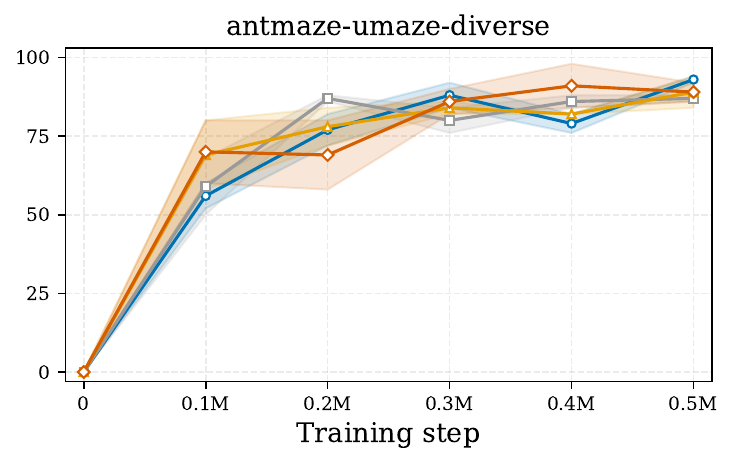}
 \end{minipage}\hfill
 \caption{\textbf{Sensitivity to the particle count $N$.} Normalized return on three D4RL AntMaze 
 tasks evaluating sensitivity to the generated particle count $N$. Curves show mean normalized returns, and
shaded regions denote one standard deviation.}
 \label{fig:particle-ablation}
\end{figure}
\begin{wrapfigure}{r}{0.42\linewidth}
    \vspace{-18pt} %
    \centering
    
    \includegraphics[width=0.9\linewidth]{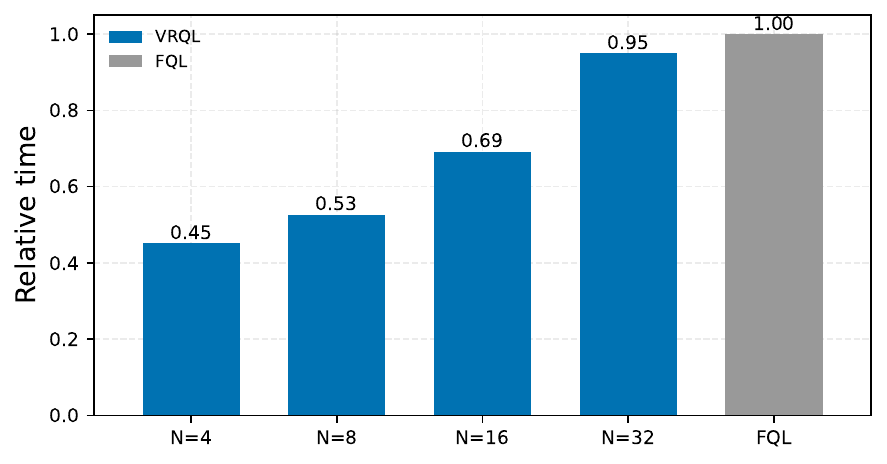}
    \caption{\textbf{Relative training time.} VRQL training time
    scales with particle count.}
    \label{fig:time-ablation}
    \vspace{-2pt} %
\end{wrapfigure}

\paragraph{Generated action particles.}
\label{sec:particle-ablation}
Figure~\ref{fig:particle-ablation} evaluates VRQL sensitivity 
to the generated particle count~$N$. 
While $N \ge 3$ is strictly necessary for value guidance to function when 
self-interactions are excluded, increasing $N$ further expands the repulsive neighborhood 
at the cost of training computation. The learning curves demonstrate that VRQL 
sensitivity to the particle count varies by environment. Specifically, 
performance on \textit{antmaze-large-play} and \textit{antmaze-medium-diverse} exhibits noticeable 
sensitivity to $N$, whereas the \textit{antmaze-umaze-diverse} task remains robust across 
all tested configurations ($N \in \{4, 8, 16, 32\}$). Concurrently, the timing 
analysis in Figure~\ref{fig:time-ablation} reveals that VRQL 
requires selecting the particle count based on specific task complexity to 
properly balance performance optimization and computational efficiency.

\FloatBarrier

\section{Related Work}
\label{sec:related-work}
\paragraph{Offline reinforcement learning.}
Offline reinforcement learning improves a policy from a fixed dataset without
 environment interaction, which makes extrapolation error on
out of distribution actions the central challenge: the critic assigns unreliable
values to actions the data never covers, and unconstrained maximization exploits
precisely those errors~\citep{levine2020offline}. One family of methods constrains
the learned policy toward the behavior policy, either by restricting the action
support through a generative model of the data or by adding an explicit divergence
penalty to the actor objective~\citep{fujimoto2019off,fujimoto2021minimalist}.
A second family regularizes the value function, penalizing estimates at
unobserved actions so that the induced policy stays
conservative~\citep{kumar2020conservative}. A third family avoids querying
out of distribution actions altogether, fitting values with in sample targets and
extracting a policy by weighted regression~\citep{kostrikov2022offline,tarasov2023revisiting}, 
and recent analyses indicate that policy extraction is often the binding constraint on
offline performance~\citep{park2024value}. VRQL adopts the policy constrained
formulation of Eq.~\ref{eq:pcorl-objective} and contributes to the regularizer, 
leaving the value estimation components unchanged.

\paragraph{Generative policies for offline reinforcement learning.}
Diffusion and flow policies provide expressive conditional action distributions
for behavioral modeling and policy extraction~\citep{florence2022implicit,janner2022planning,wang2023diffusion,hansenestruch2023idql,park2025flow}. A first line of work applies
value improvement through an iterative sampler, which can require differentiating
through multiple denoising or flow integration steps~\citep{wang2023diffusion,lu2023contrastive,doo2026qflow}. A second line first fits an expressive
behavior model and then uses its samples or predictions to regularize a
separately trained policy~\citep{chen2023offline,hansenestruch2023idql,park2025flow}. This two stage design separates the
policy objective from the behavior model, but introduces an additional model
and an approximation error between the two stages. VRQL instead modifies
the interactions among actions generated by the policy regularizer itself,
within a direct policy training objective. Sampling cost has also been reduced by distilled or few step policy
classes~\citep{kang2023efficient,ding2024consistency}, and entropy or ensemble
regularization has been used to preserve action diversity~\citep{zhang2024entropy}.

\paragraph{Value-aware behavioral regularization.}
Offline methods with value-aware behavioral regularization use estimated action quality to make
behavioral fitting selective. Advantage weighted regression and actor-critic
policy extraction methods emphasize behavior action with higher estimated
advantage, while implicit value based extraction combines in sample value
learning with a similarly selective fitting objective~\citep{peters2007reinforcement,peng2019advantage,nair2020awac,kostrikov2022offline}. 
Value-aware fitting has also been applied to expressive
generative policies, where value or advantage estimates reweight flow matching
or diffusion targets~\citep{ajay2023conditional,ding2024diffusion,zhang2025energyweighted}. 
Auxiliary behavior model methods can use the same signal when
fitting a behavioral model and then transfer that model's samples or targets
to a separate actor~\citep{tiofack2026guided}.
These approaches establish that behavioral regularization need not treat all
examples equally, but they place the value weight on behavior action,
training examples, or model targets. Value-aware fitting methods use critic
estimates to prioritize behavioral examples or auxiliary model targets; VRQL
uses critic estimates to assign interaction weights to the generated neighbors
that define repulsion.

\paragraph{Drifting models and policies.}
Drifting models train a direct generator by regressing toward stopped targets
constructed from attractive and repulsive sample interactions~\citep{deng2026generative}. 
A conditional offline policy construction specializes this principle
to action generation by combining attraction toward a behavior action,
repulsion among generated actions, and a direct actor value objective~\citep{houssaini2026drift}. 
VRQL retains this geometric
foundation and introduces a negative value dependent bias into the normalized
repulsive weights. Kernel similarity determines geometric influence, while
relative critic values determine which generated neighbors exert more or less
of that influence. The resulting contribution is a value-aware interaction rule
inside the drifting regularizer.

\section{Conclusion}
\label{sec:conclusion}

We introduced VRQL, an offline algorithm that incorporates estimated action values 
directly into the repulsive interactions of a drifting policy. By combining geometric 
similarity with value dependent weights, VRQL modulates the influence of generated 
neighbors while preserving attraction toward behavioral data and maintaining a 
separate value maximization objective. Furthermore, batch level scaling effectively 
calibrates this guidance against typical within state value variations. Empirical 
evaluations across the OGBench and D4RL benchmarks demonstrate that VRQL establishes 
strong performance and outperforms representative baselines. Ablation studies suggest 
that these empirical gains are associated with pairing low action values with high 
geometric repulsion. Disrupting this learned assignment severely degrades policy 
optimization, whereas proper within state calibration and moderate guidance temperatures 
ensure stable learning.

However, a primary limitation of this approach stems from its inherent reliance on learned 
critic estimates. In environments where the critic produces incorrect value orderings, 
the guidance mechanism can inadvertently repel optimal actions and misguide the policy 
constraint. Future work will investigate integrating uncertainty estimation into the particle 
interaction field, allowing the algorithm to dynamically attenuate value guidance when critic 
reliability degrades. Additionally, future research will explore extending this value-aware
mechanism to the behavioral attraction phase, allowing the policy to preferentially anchor 
to high return data points.

\clearpage
\subsection*{Reproducibility statement}
To support reproducibility, the main paper describes the method and training
objective, Appendix~\ref{app:experimental-details} provides the network
architectures, hyperparameters, benchmark definitions, and compute resources,
and the supplementary materials contain anonymized source code.

\subsection*{AI use statement}
AI tools were used solely to improve the language and readability of
this article. They were not used to generate the method, experimental results,
scientific claims, or citations. All AI-assisted edits were reviewed by the
authors, who take full responsibility for the final content of this work.

\bibliography{references}
\bibliographystyle{iclr2027_conference}

\clearpage
\appendix

\section{Implementation Details}
\label{app:experimental-details}

This section describes the implementation of VRQL, including the network
architectures, hyperparameter settings, and benchmark environments used in the
experiments.

\subsection{Network Architectures}
\label{app:network-architectures}

We parameterize the policy generator $f_\theta$ and the critic ensemble
$\{Q_{\phi_h}\}_{h=1}^{H}$ using multilayer perceptrons (MLPs). The actor and
critics have separate parameters and are trained with Adam.

\paragraph{Actor network ($f_\theta$).}
The actor consists of four hidden layers with $512$ units each and GELU
activations, without layer normalization. Its inputs are the state $s$ and a
Gaussian noise vector $z\sim\mathcal N(0,I)$ whose dimension equals the action
dimension. The output is clipped coordinatewise to the action box
$\mathcal A=[-1,1]^d$, giving $x_\theta(s,z)=\Pi_{\mathcal A}(f_\theta(s,z))$.
Sampling $z$ induces the policy $\pi_\theta(\cdot\mid s)$.

\paragraph{Critic networks ($Q_{\phi_h}$).}
We use an ensemble of $H=2$ critics. Each critic takes a state and an action as
input and consists of four hidden layers with $512$ units each, GELU
activations, and layer normalization. The actor objective uses the mean of
the online critic estimates, whereas Bellman targets use the minimum of the
target critic estimates, as specified in Eqs.~\ref{eq:actor-objective}
and~\ref{eq:critic-target}. By default, particle value guidance also uses the
minimum of the target critic estimates. Target parameters are updated by
Polyak averaging with coefficient $\rho$.

\subsection{Hyperparameters}
\label{app:hyperparameters}
\label{app:ablation-details}
\label{app:normalization-details}

The shared training settings for VRQL are summarized in
Table~\ref{tab:shared-hparams}, and the value guidance settings are listed in
Table~\ref{tab:repulsion-hparams}. All runs use the Laplace kernel defined in
Eq.~\ref{eq:kernels}. We use $\tau$ for the kernel temperature and $\rho$ for
the target network update coefficient.

\begin{table}[h]
\caption{Shared hyperparameters. Symbols refer to the main text where one is defined.}
\label{tab:shared-hparams}

\vspace{\baselineskip}
\centering
\begin{tabular}{llr}
\toprule
\textbf{Hyperparameter} & \textbf{Symbol} & \textbf{Value} \\
\midrule
\multicolumn{3}{c}{\textit{General training settings}} \\
\midrule
Optimizer & -- & Adam \\
Actor / critic learning rate & -- & $3\times10^{-4}$ \\
Batch size & -- & $256$ \\
Hidden dimensions (actor and critic) & -- & $(512,512,512,512)$ \\
Activation & -- & GELU \\
Critic ensemble size & $H$ & $2$ \\
Target update coefficient & $\rho$ & $0.005$ \\
\midrule
\multicolumn{3}{c}{\textit{Drifting policy settings}} \\
\midrule
Particle count & $N$ & $32$ \\
Drift step size & $\eta$ & $1.0$ \\
Drift batch size & $B_d$ & $256$ \\
Drift distance stabilizer & -- & $10^{-12}$ \\
\midrule
\multicolumn{3}{c}{\textit{Training duration}} \\
\midrule
Offline gradient steps (OGBench) & -- & $1{,}000{,}000$ \\
Offline gradient steps (D4RL AntMaze) & -- & $500{,}000$ \\
\bottomrule
\end{tabular}
\end{table}

\begin{table}[h]
\caption{Value-guided repulsion settings. These are the defaults of
Section~\ref{sec:value-repulsion}; the ablations of Section~\ref{sec:ablations}
vary them one at a time.}
\label{tab:repulsion-hparams}
\vspace{\baselineskip}
\centering
\begin{tabular}{lll}
\toprule
\textbf{Hyperparameter} & \textbf{Default} & \textbf{Alternatives} \\
\midrule
Value scale $S_{\mathrm{norm}}$ & $S_{\mathrm{std}}$ (Eq.~\ref{eq:value-scale})
  & $S_{\mathrm{abs}}$, $S_{\mathrm{none}}$ (Eq.~\ref{eq:normalization-ablation}) \\
Ensemble aggregation $\operatorname{Agg}$ & $\min$ & mean \\
Critic for the value bias & target & online \\
Value scale stabilizer $\epsilon$ & $10^{-6}$ & -- \\
\bottomrule
\end{tabular}
\end{table}

\paragraph{Task specific hyperparameters.}
Following the hyperparameter protocol of DriftQL~\citep{houssaini2026drift},
we retain its general training conventions and adapt the settings needed for
value-guided repulsion.
The behavioral regularization coefficient $\alpha$ controls the balance between
the drift regression loss and direct value maximization in
Eq.~\ref{eq:actor-objective}. We tune the guidance temperature $\beta$ and
the kernel temperature $\tau$ by task.
Across OGBench, the settings span $\beta\in[0.05,10]$,
and $\tau\in[0.02,0.8]$. D4RL AntMaze uses $\beta=1$. The per task
configurations are released with the code.

\subsection{Environment Details}
\label{app:environment-details}
\label{app:task-catalog}

We evaluate VRQL on $56$ task configurations: $50$ from OGBench and $6$ from
D4RL AntMaze. All experiments train on fixed offline datasets.

\paragraph{OGBench.}
The benchmark covers navigation, object manipulation, and puzzle solving.
Navigation tasks include \textit{antmaze-large-navigate},
\textit{antmaze-giant-navigate}, \textit{humanoidmaze-medium-navigate},
\textit{humanoidmaze-large-navigate}, and \textit{antsoccer-arena-navigate}.
Manipulation and puzzle tasks include \textit{cube-single-play},
\textit{cube-double-play}, \textit{scene-play}, \textit{puzzle-3x3-play}, and
\textit{puzzle-4x4-play}. Each of these ten families has five single task
configurations. The full identifier appends
\textit{-singletask-task}$i$\textit{-v0}, for $i\in\{1,2,3,4,5\}$, to the
family prefix. Table~\ref{tab:main_results} abbreviates these prefixes by
omitting the \textit{-navigate} or \textit{-play} suffix.

\paragraph{D4RL AntMaze.}
The six configurations cover three maze sizes and different dataset variants:
\textit{antmaze-umaze-v2}, \textit{antmaze-umaze-diverse-v2},
\textit{antmaze-medium-play-v2}, \textit{antmaze-medium-diverse-v2},
\textit{antmaze-large-play-v2}, and \textit{antmaze-large-diverse-v2}.
These tasks evaluate navigation using offline trajectories with different
maze layouts and dataset compositions.

Both benchmarks use a bounded action box $\mathcal A=[-1,1]^d$, so the projection
$\Pi_{\mathcal A}$ is a coordinatewise clip. We report success rate on OGBench
and normalized return on D4RL AntMaze, following the protocol of
Section~\ref{sec:experimental-setup} and the conventions of the respective
benchmarks~\citep{fu2020d4rl,park2025ogbench}.

\section{Compute Resources}
\label{app:compute-resources}

All experiments run on a machine equipped with an AMD 9950 CPU and a single
NVIDIA RTX 5090 GPU, with one task per device, in a JAX implementation. A single
OGBench run takes approximately $0.5$ GPU hours for
$1{,}000{,}000$ gradient steps, and a single D4RL AntMaze run approximately
$0.2$ GPU hours for $500{,}000$ steps; both figures include periodic policy
evaluation. With five seeds per configuration, the $50$ OGBench and 6 D4RL
configurations of Table~\ref{tab:main_results} amount to roughly $125$ and
$6$ GPU hours respectively, excluding the ablations of
Section~\ref{sec:ablations} and the hyperparameter search.

\section{Detailed benchmark results}
\label{app:benchmark-results}

The tables below give the complete baseline inventory and the per task results
underlying the main comparison. OGBench rows are grouped by the identifier
prefixes of Appendix~\ref{app:task-catalog} and named by their
\textit{task}$i$\textit{-v0} suffix; D4RL rows are the 6 AntMaze v2 tasks
listed above, named by their dataset qualifier (\textit{div} abbreviates
\textit{diverse}). The fourteen methods are split across two tables, seven
each.

\begingroup
\setlength{\tabcolsep}{5pt}
\begin{longtable}{@{}l *{7}{r}@{}}
\caption{Full results across all evaluated OGBench and D4RL tasks, methods one to seven,  
Boldface marks the best of all fourteen methods on a task, a row's best may appear in
Table~\ref{tab:all_results_2}, which gives the remaining seven methods.}
\label{tab:all_results} \\
\toprule
& \multicolumn{3}{c}{Gaussian} & \multicolumn{3}{c}{Diffusion} & \multicolumn{1}{c}{Flow} \\
\cmidrule(lr){2-4} \cmidrule(lr){5-7} \cmidrule(lr){8-8}
Task & BC & IQL & ReBRAC & IDQL & SRPO & CAC & FAWAC \\
\midrule
\endfirsthead
\multicolumn{8}{@{}l}{Table~\thetable{} (continued)} \\
\toprule
& \multicolumn{3}{c}{Gaussian} & \multicolumn{3}{c}{Diffusion} & \multicolumn{1}{c}{Flow} \\
\cmidrule(lr){2-4} \cmidrule(lr){5-7} \cmidrule(lr){8-8}
Task & BC & IQL & ReBRAC & IDQL & SRPO & CAC & FAWAC \\
\midrule
\endhead
\midrule
\multicolumn{8}{r@{}}{Continued on next page} \\
\endfoot
\bottomrule
\endlastfoot
\multicolumn{8}{@{}l}{\textit{antmaze-large-navigate-singletask}} \\*
task1-v0 $\star$ & $0 \pm 0$ & $48 \pm 9$ & $91 \pm 10$ & $0 \pm 0$ & $0 \pm 0$ & $42 \pm 7$ & $1 \pm 1$ \\*
task2-v0 & $6 \pm 3$ & $42 \pm 6$ & $88 \pm 4$ & $14 \pm 8$ & $4 \pm 4$ & $1 \pm 1$ & $0 \pm 1$ \\*
task3-v0 & $29 \pm 5$ & $72 \pm 7$ & $51 \pm 18$ & $26 \pm 8$ & $3 \pm 2$ & $49 \pm 10$ & $12 \pm 4$ \\*
task4-v0 & $8 \pm 3$ & $51 \pm 9$ & $84 \pm 7$ & $62 \pm 25$ & $45 \pm 19$ & $17 \pm 6$ & $10 \pm 3$ \\*
task5-v0 & $10 \pm 3$ & $54 \pm 22$ & $90 \pm 2$ & $2 \pm 2$ & $1 \pm 1$ & $55 \pm 6$ & $9 \pm 5$ \\
\midrule
\multicolumn{8}{@{}l}{\textit{antmaze-giant-navigate-singletask}} \\*
task1-v0 $\star$ & $0 \pm 0$ & $0 \pm 0$ & $27 \pm 22$ & $0 \pm 0$ & $0 \pm 0$ & $0 \pm 0$ & $0 \pm 0$ \\*
task2-v0 & $0 \pm 0$ & $1 \pm 1$ & $16 \pm 17$ & $0 \pm 0$ & $0 \pm 0$ & $0 \pm 0$ & $0 \pm 0$ \\*
task3-v0 & $0 \pm 0$ & $0 \pm 0$ & $34 \pm 22$ & $0 \pm 0$ & $0 \pm 0$ & $0 \pm 0$ & $0 \pm 0$ \\*
task4-v0 & $0 \pm 0$ & $0 \pm 0$ & $5 \pm 12$ & $0 \pm 0$ & $0 \pm 0$ & $0 \pm 0$ & $0 \pm 0$ \\*
task5-v0 & $1 \pm 1$ & $19 \pm 7$ & $49 \pm 22$ & $0 \pm 1$ & $0 \pm 0$ & $0 \pm 0$ & $0 \pm 0$ \\
\midrule
\multicolumn{8}{@{}l}{\textit{humanoidmaze-medium-navigate-singletask}} \\*
task1-v0 $\star$ & $1 \pm 0$ & $32 \pm 7$ & $16 \pm 9$ & $1 \pm 1$ & $0 \pm 0$ & $38 \pm 19$ & $6 \pm 2$ \\*
task2-v0 & $1 \pm 0$ & $41 \pm 9$ & $18 \pm 16$ & $1 \pm 1$ & $1 \pm 1$ & $47 \pm 35$ & $40 \pm 2$ \\*
task3-v0 & $6 \pm 2$ & $25 \pm 6$ & $36 \pm 13$ & $0 \pm 1$ & $2 \pm 1$ & $83 \pm 18$ & $19 \pm 2$ \\*
task4-v0 & $0 \pm 0$ & $0 \pm 1$ & $15 \pm 16$ & $1 \pm 1$ & $1 \pm 1$ & $5 \pm 4$ & $1 \pm 1$ \\*
task5-v0 & $2 \pm 1$ & $66 \pm 4$ & $24 \pm 20$ & $1 \pm 1$ & $3 \pm 3$ & $91 \pm 5$ & $31 \pm 7$ \\
\midrule
\multicolumn{8}{@{}l}{\textit{humanoidmaze-large-navigate-singletask}} \\*
task1-v0 $\star$ & $0 \pm 0$ & $3 \pm 1$ & $2 \pm 1$ & $0 \pm 0$ & $0 \pm 0$ & $1 \pm 1$ & $0 \pm 0$ \\*
task2-v0 & $0 \pm 0$ & $0 \pm 0$ & $0 \pm 0$ & $0 \pm 0$ & $0 \pm 0$ & $0 \pm 0$ & $0 \pm 0$ \\*
task3-v0 & $1 \pm 1$ & $7 \pm 3$ & $8 \pm 4$ & $3 \pm 1$ & $1 \pm 1$ & $2 \pm 3$ & $1 \pm 1$ \\*
task4-v0 & $1 \pm 0$ & $1 \pm 0$ & $1 \pm 1$ & $0 \pm 0$ & $0 \pm 0$ & $0 \pm 1$ & $0 \pm 0$ \\*
task5-v0 & $0 \pm 1$ & $1 \pm 1$ & $2 \pm 2$ & $0 \pm 0$ & $0 \pm 0$ & $0 \pm 0$ & $0 \pm 0$ \\
\midrule
\multicolumn{8}{@{}l}{\textit{antsoccer-arena-navigate-singletask}} \\*
task1-v0 & $2 \pm 1$ & $14 \pm 6$ & $0 \pm 0$ & $44 \pm 12$ & $2 \pm 1$ & $1 \pm 3$ & $22 \pm 2$ \\*
task2-v0 & $2 \pm 2$ & $17 \pm 7$ & $0 \pm 1$ & $15 \pm 12$ & $3 \pm 1$ & $0 \pm 0$ & $8 \pm 1$ \\*
task3-v0 & $0 \pm 0$ & $6 \pm 4$ & $0 \pm 0$ & $0 \pm 0$ & $0 \pm 0$ & $8 \pm 19$ & $11 \pm 5$ \\*
task4-v0 $\star$ & $1 \pm 0$ & $3 \pm 2$ & $0 \pm 0$ & $0 \pm 1$ & $0 \pm 0$ & $0 \pm 0$ & $12 \pm 3$ \\*
task5-v0 & $0 \pm 0$ & $2 \pm 2$ & $0 \pm 0$ & $0 \pm 0$ & $0 \pm 0$ & $0 \pm 0$ & $9 \pm 2$ \\
\midrule
\multicolumn{8}{@{}l}{\textit{cube-single-play-singletask}} \\*
task1-v0 & $10 \pm 5$ & $88 \pm 3$ & $89 \pm 5$ & $95 \pm 2$ & $89 \pm 7$ & $77 \pm 28$ & $81 \pm 9$ \\*
task2-v0 $\star$ & $3 \pm 1$ & $85 \pm 8$ & $92 \pm 4$ & $96 \pm 2$ & $82 \pm 16$ & $80 \pm 30$ & $81 \pm 9$ \\*
task3-v0 & $9 \pm 3$ & $91 \pm 5$ & $93 \pm 3$ & $99 \pm 1$ & $96 \pm 2$ & $98 \pm 1$ & $87 \pm 4$ \\*
task4-v0 & $2 \pm 1$ & $73 \pm 6$ & $92 \pm 3$ & $93 \pm 4$ & $70 \pm 18$ & $91 \pm 2$ & $79 \pm 6$ \\*
task5-v0 & $3 \pm 3$ & $78 \pm 8$ & $90 \pm 8$ & $90 \pm 6$ & $61 \pm 12$ & $80 \pm 20$ & $78 \pm 10$ \\
\midrule
\multicolumn{8}{@{}l}{\textit{cube-double-play-singletask}} \\*
task1-v0 & $8 \pm 3$ & $27 \pm 5$ & $45 \pm 6$ & $39 \pm 19$ & $7 \pm 6$ & $21 \pm 8$ & $21 \pm 7$ \\*
task2-v0 $\star$ & $0 \pm 0$ & $1 \pm 1$ & $7 \pm 3$ & $16 \pm 10$ & $0 \pm 0$ & $2 \pm 2$ & $2 \pm 1$ \\*
task3-v0 & $0 \pm 0$ & $0 \pm 0$ & $4 \pm 1$ & $17 \pm 8$ & $0 \pm 1$ & $3 \pm 1$ & $1 \pm 1$ \\*
task4-v0 & $0 \pm 0$ & $0 \pm 0$ & $1 \pm 1$ & $0 \pm 1$ & $0 \pm 0$ & $0 \pm 0$ & $0 \pm 0$ \\*
task5-v0 & $0 \pm 0$ & $4 \pm 3$ & $4 \pm 2$ & $1 \pm 1$ & $0 \pm 0$ & $3 \pm 2$ & $2 \pm 1$ \\
\midrule
\multicolumn{8}{@{}l}{\textit{scene-play-singletask}} \\*
task1-v0 & $19 \pm 6$ & $94 \pm 3$ & $95 \pm 2$ & $\textbf{100} \pm 0$ & $94 \pm 4$ & $\textbf{100} \pm 1$ & $87 \pm 8$ \\*
task2-v0 $\star$ & $1 \pm 1$ & $12 \pm 3$ & $50 \pm 13$ & $33 \pm 14$ & $2 \pm 2$ & $50 \pm 40$ & $18 \pm 8$ \\*
task3-v0 & $1 \pm 1$ & $32 \pm 7$ & $55 \pm 16$ & $94 \pm 4$ & $4 \pm 4$ & $49 \pm 16$ & $38 \pm 9$ \\*
task4-v0 & $2 \pm 2$ & $0 \pm 1$ & $3 \pm 3$ & $4 \pm 3$ & $0 \pm 0$ & $0 \pm 0$ & $6 \pm 1$ \\*
task5-v0 & $0 \pm 0$ & $0 \pm 0$ & $0 \pm 0$ & $0 \pm 0$ & $0 \pm 0$ & $0 \pm 0$ & $0 \pm 0$ \\
\midrule
\multicolumn{8}{@{}l}{\textit{puzzle-3x3-play-singletask}} \\*
task1-v0 & $5 \pm 2$ & $33 \pm 6$ & $97 \pm 4$ & $52 \pm 12$ & $89 \pm 5$ & $97 \pm 2$ & $25 \pm 9$ \\*
task2-v0 & $1 \pm 1$ & $4 \pm 3$ & $1 \pm 1$ & $0 \pm 1$ & $0 \pm 1$ & $0 \pm 0$ & $4 \pm 2$ \\*
task3-v0 & $1 \pm 1$ & $3 \pm 2$ & $3 \pm 1$ & $0 \pm 0$ & $0 \pm 0$ & $0 \pm 0$ & $1 \pm 1$ \\*
task4-v0 $\star$ & $1 \pm 1$ & $2 \pm 1$ & $2 \pm 1$ & $0 \pm 0$ & $0 \pm 0$ & $0 \pm 0$ & $1 \pm 1$ \\*
task5-v0 & $1 \pm 0$ & $3 \pm 2$ & $5 \pm 3$ & $0 \pm 0$ & $0 \pm 0$ & $0 \pm 0$ & $1 \pm 1$ \\
\midrule
\multicolumn{8}{@{}l}{\textit{puzzle-4x4-play-singletask}} \\*
task1-v0 & $1 \pm 1$ & $12 \pm 2$ & $26 \pm 4$ & $48 \pm 5$ & $24 \pm 9$ & $44 \pm 10$ & $1 \pm 2$ \\*
task2-v0 & $0 \pm 0$ & $7 \pm 4$ & $12 \pm 4$ & $14 \pm 5$ & $0 \pm 1$ & $0 \pm 0$ & $0 \pm 1$ \\*
task3-v0 & $0 \pm 0$ & $9 \pm 3$ & $15 \pm 3$ & $34 \pm 5$ & $21 \pm 10$ & $29 \pm 12$ & $1 \pm 1$ \\*
task4-v0 $\star$ & $0 \pm 0$ & $5 \pm 2$ & $10 \pm 3$ & $\textbf{26} \pm 6$ & $7 \pm 4$ & $1 \pm 1$ & $0 \pm 0$ \\*
task5-v0 & $0 \pm 0$ & $4 \pm 1$ & $7 \pm 3$ & $\textbf{24} \pm 11$ & $1 \pm 1$ & $0 \pm 0$ & $0 \pm 1$ \\
\midrule
\multicolumn{8}{@{}l}{\textit{antmaze-$\ast$-v2 (D4RL)}} \\*
umaze & $55$ & $77$ & $98$ & $94$ & $97$ & $66 \pm 5$ & $90 \pm 6$ \\*
umaze-div & $47$ & $54$ & $84$ & $80$ & $82$ & $66 \pm 11$ & $55 \pm 7$ \\*
medium-play & $0$ & $66$ & $\textbf{90}$ & $84$ & $81$ & $49 \pm 24$ & $52 \pm 12$ \\*
medium-div & $1$ & $74$ & $84$ & $\textbf{85}$ & $75$ & $0 \pm 1$ & $44 \pm 15$ \\*
large-play & $0$ & $42$ & $52$ & $64$ & $54$ & $0 \pm 0$ & $10 \pm 6$ \\*
large-div & $0$ & $30$ & $64$ & $68$ & $54$ & $0 \pm 0$ & $16 \pm 10$ \\
\end{longtable}
\endgroup
\clearpage
\begingroup
\setlength{\tabcolsep}{5pt}
\begin{longtable}{@{}l *{7}{r}@{}}
\caption{Full results across all evaluated OGBench and D4RL tasks, methods eight to fourteen, continuing
Table~\ref{tab:all_results} under the same conventions; boldface again marks
the best of all fourteen methods on a task.}
\label{tab:all_results_2} \\
\toprule
& \multicolumn{3}{c}{Flow} & \multicolumn{2}{c}{Value-Guide} & \multicolumn{2}{c}{Drift} \\
\cmidrule(lr){2-4} \cmidrule(lr){5-6} \cmidrule(lr){7-8}
Task & FBRAC & IFQL & FQL & Q-Flow & GFP & DriftQL & \textbf{VRQL} \\
\midrule
\endfirsthead
\multicolumn{8}{@{}l}{Table~\thetable{} (continued)} \\
\toprule
& \multicolumn{3}{c}{Flow} & \multicolumn{2}{c}{Value-Guide} & \multicolumn{2}{c}{Drift} \\
\cmidrule(lr){2-4} \cmidrule(lr){5-6} \cmidrule(lr){7-8}
Task & FBRAC & IFQL & FQL & Q-Flow & GFP & DriftQL & \textbf{VRQL} \\
\midrule
\endhead
\midrule
\multicolumn{8}{r@{}}{Continued on next page} \\
\endfoot
\bottomrule
\endlastfoot
\multicolumn{8}{@{}l}{\textit{antmaze-large-navigate-singletask}} \\*
task1-v0 $\star$ & $70 \pm 20$ & $24 \pm 17$ & $80 \pm 8$ & $95 \pm 4$ & $95 \pm 0$ & $95 \pm 2$ & $\textbf{99} \pm 1$ \\*
task2-v0 & $35 \pm 12$ & $8 \pm 3$ & $57 \pm 10$ & $85 \pm 6$ & $\textbf{92} \pm 3$ & $85 \pm 9$ & $91 \pm 0$ \\*
task3-v0 & $83 \pm 15$ & $52 \pm 17$ & $93 \pm 3$ & $93 \pm 4$ & $95 \pm 2$ & $97 \pm 1$ & $\textbf{99} \pm 1$ \\*
task4-v0 & $37 \pm 18$ & $18 \pm 8$ & $80 \pm 4$ & $79 \pm 23$ & $90 \pm 2$ & $91 \pm 1$ & $\textbf{97} \pm 2$ \\*
task5-v0 & $76 \pm 8$ & $38 \pm 18$ & $83 \pm 4$ & $90 \pm 6$ & $95 \pm 1$ & $92 \pm 1$ & $\textbf{97} \pm 1$ \\
\midrule
\multicolumn{8}{@{}l}{\textit{antmaze-giant-navigate-singletask}} \\*
task1-v0 $\star$ & $0 \pm 1$ & $0 \pm 0$ & $4 \pm 5$ & $15 \pm 10$ & $12 \pm 15$ & $32 \pm 2$ & $\textbf{56} \pm 14$ \\*
task2-v0 & $4 \pm 7$ & $0 \pm 0$ & $9 \pm 7$ & $34 \pm 22$ & $52 \pm 26$ & $\textbf{79} \pm 1$ & $69 \pm 9$ \\*
task3-v0 & $0 \pm 0$ & $0 \pm 0$ & $0 \pm 1$ & $9 \pm 5$ & $13 \pm 10$ & $43 \pm 2$ & $\textbf{67} \pm 7$ \\*
task4-v0 & $9 \pm 4$ & $0 \pm 0$ & $14 \pm 23$ & $68 \pm 13$ & $17 \pm 19$ & $64 \pm 3$ & $\textbf{87} \pm 4$ \\*
task5-v0 & $6 \pm 10$ & $13 \pm 9$ & $16 \pm 28$ & $73 \pm 12$ & $43 \pm 35$ & $85 \pm 5$ & $\textbf{89} \pm 3$ \\
\midrule
\multicolumn{8}{@{}l}{\textit{humanoidmaze-medium-navigate-singletask}} \\*
task1-v0 $\star$ & $25 \pm 8$ & $69 \pm 19$ & $19 \pm 12$ & $\textbf{87} \pm 5$ & $83 \pm 3$ & $28 \pm 2$ & $\textbf{87} \pm 12$ \\*
task2-v0 & $76 \pm 10$ & $85 \pm 11$ & $94 \pm 3$ & $95 \pm 4$ & $91 \pm 6$ & $87 \pm 2$ & $\textbf{100} \pm 0$ \\*
task3-v0 & $27 \pm 11$ & $49 \pm 49$ & $74 \pm 18$ & $95 \pm 3$ & $86 \pm 10$ & $56 \pm 3$ & $\textbf{99} \pm 2$ \\*
task4-v0 & $1 \pm 2$ & $1 \pm 1$ & $3 \pm 4$ & $39 \pm 21$ & $3 \pm 6$ & $39 \pm 3$ & $\textbf{56} \pm 37$ \\*
task5-v0 & $63 \pm 9$ & $98 \pm 2$ & $97 \pm 2$ & $98 \pm 2$ & $95 \pm 1$ & $99 \pm 2$ & $\textbf{100} \pm 0$ \\
\midrule
\multicolumn{8}{@{}l}{\textit{humanoidmaze-large-navigate-singletask}} \\*
task1-v0 $\star$ & $0 \pm 0$ & $6 \pm 2$ & $7 \pm 6$ & $14 \pm 7$ & $\textbf{57} \pm 23$ & $2 \pm 1$ & $13 \pm 3$ \\*
task2-v0 & $0 \pm 0$ & $0 \pm 0$ & $0 \pm 0$ & $0 \pm 1$ & $0 \pm 0$ & $0 \pm 0$ & $0 \pm 0$ \\*
task3-v0 & $10 \pm 2$ & $\textbf{48} \pm 10$ & $11 \pm 7$ & $16 \pm 5$ & $14 \pm 16$ & $23 \pm 1$ & $24 \pm 1$ \\*
task4-v0 & $1 \pm 1$ & $1 \pm 1$ & $2 \pm 3$ & $\textbf{5} \pm 5$ & $3 \pm 4$ & $1 \pm 1$ & $0 \pm 1$ \\*
task5-v0 & $1 \pm 1$ & $0 \pm 0$ & $1 \pm 3$ & $\textbf{5} \pm 4$ & $3 \pm 15$ & $1 \pm 0$ & $1 \pm 1$ \\
\midrule
\multicolumn{8}{@{}l}{\textit{antsoccer-arena-navigate-singletask}} \\*
task1-v0 & $17 \pm 3$ & $61 \pm 25$ & $77 \pm 4$ & $73 \pm 9$ & $77 \pm 1$ & $79 \pm 4$ & $\textbf{81} \pm 2$ \\*
task2-v0 & $8 \pm 2$ & $75 \pm 3$ & $88 \pm 3$ & $94 \pm 5$ & $91 \pm 2$ & $91 \pm 2$ & $\textbf{98} \pm 3$ \\*
task3-v0 & $16 \pm 3$ & $14 \pm 22$ & $61 \pm 6$ & $58 \pm 9$ & $51 \pm 4$ & $60 \pm 3$ & $\textbf{66} \pm 1$ \\*
task4-v0 $\star$ & $24 \pm 4$ & $16 \pm 9$ & $39 \pm 6$ & $27 \pm 11$ & $40 \pm 4$ & $48 \pm 5$ & $\textbf{50} \pm 1$ \\*
task5-v0 & $15 \pm 4$ & $0 \pm 1$ & $36 \pm 9$ & $25 \pm 15$ & $29 \pm 8$ & $48 \pm 9$ & $\textbf{58} \pm 8$ \\
\midrule
\multicolumn{8}{@{}l}{\textit{cube-single-play-singletask}} \\*
task1-v0 & $73 \pm 33$ & $79 \pm 4$ & $97 \pm 2$ & $95 \pm 3$ & $99 \pm 0$ & $94 \pm 3$ & $\textbf{100} \pm 1$ \\*
task2-v0 $\star$ & $83 \pm 13$ & $73 \pm 3$ & $97 \pm 2$ & $98 \pm 2$ & $\textbf{99} \pm 0$ & $93 \pm 2$ & $\textbf{99} \pm 1$ \\*
task3-v0 & $82 \pm 12$ & $88 \pm 4$ & $98 \pm 2$ & $98 \pm 2$ & $99 \pm 0$ & $95 \pm 2$ & $\textbf{100} \pm 0$ \\*
task4-v0 & $79 \pm 20$ & $79 \pm 6$ & $94 \pm 3$ & $90 \pm 5$ & $\textbf{99} \pm 0$ & $92 \pm 3$ & $97 \pm 2$ \\*
task5-v0 & $76 \pm 33$ & $77 \pm 7$ & $93 \pm 3$ & $92 \pm 5$ & $97 \pm 1$ & $90 \pm 3$ & $\textbf{98} \pm 1$ \\
\midrule
\multicolumn{8}{@{}l}{\textit{cube-double-play-singletask}} \\*
task1-v0 & $47 \pm 11$ & $35 \pm 9$ & $61 \pm 9$ & $67 \pm 11$ & $76 \pm 4$ & $49 \pm 1$ & $\textbf{95} \pm 3$ \\*
task2-v0 $\star$ & $22 \pm 12$ & $9 \pm 5$ & $36 \pm 6$ & $27 \pm 10$ & $53 \pm 8$ & $23 \pm 5$ & $\textbf{68} \pm 1$ \\*
task3-v0 & $4 \pm 2$ & $8 \pm 5$ & $22 \pm 5$ & $28 \pm 10$ & $43 \pm 8$ & $9 \pm 3$ & $\textbf{58} \pm 11$ \\*
task4-v0 & $0 \pm 1$ & $1 \pm 1$ & $5 \pm 2$ & $\textbf{9} \pm 5$ & $7 \pm 3$ & $3 \pm 1$ & $\textbf{9} \pm 1$ \\*
task5-v0 & $2 \pm 2$ & $17 \pm 6$ & $19 \pm 10$ & $48 \pm 20$ & $56 \pm 11$ & $43 \pm 2$ & $\textbf{60} \pm 10$ \\
\midrule
\multicolumn{8}{@{}l}{\textit{scene-play-singletask}} \\*
task1-v0 & $96 \pm 8$ & $98 \pm 3$ & $\textbf{100} \pm 0$ & $\textbf{100} \pm 0$ & $99 \pm 0$ & $\textbf{100} \pm 0$ & $\textbf{100} \pm 0$ \\*
task2-v0 $\star$ & $46 \pm 10$ & $0 \pm 0$ & $76 \pm 9$ & $\textbf{96} \pm 6$ & $89 \pm 4$ & $89 \pm 3$ & $91 \pm 3$ \\*
task3-v0 & $78 \pm 14$ & $54 \pm 19$ & $\textbf{98} \pm 1$ & $96 \pm 4$ & $78 \pm 13$ & $93 \pm 3$ & $95 \pm 5$ \\*
task4-v0 & $4 \pm 4$ & $0 \pm 0$ & $5 \pm 1$ & $7 \pm 9$ & $0 \pm 0$ & $83 \pm 10$ & $\textbf{87} \pm 2$ \\*
task5-v0 & $0 \pm 0$ & $0 \pm 0$ & $0 \pm 0$ & $0 \pm 1$ & $0 \pm 0$ & $\textbf{2} \pm 2$ & $\textbf{2} \pm 2$ \\
\midrule
\multicolumn{8}{@{}l}{\textit{puzzle-3x3-play-singletask}} \\*
task1-v0 & $63 \pm 19$ & $94 \pm 3$ & $90 \pm 4$ & $94 \pm 4$ & $94 \pm 4$ & $87 \pm 9$ & $\textbf{100} \pm 0$ \\*
task2-v0 & $2 \pm 2$ & $1 \pm 2$ & $16 \pm 5$ & $\textbf{60} \pm 17$ & $0 \pm 0$ & $39 \pm 2$ & $10 \pm 5$ \\*
task3-v0 & $1 \pm 1$ & $0 \pm 0$ & $10 \pm 3$ & $\textbf{28} \pm 7$ & $0 \pm 0$ & $20 \pm 3$ & $6 \pm 0$ \\*
task4-v0 $\star$ & $2 \pm 2$ & $0 \pm 0$ & $16 \pm 5$ & $35 \pm 9$ & $5 \pm 2$ & $10 \pm 6$ & $\textbf{84} \pm 2$ \\*
task5-v0 & $2 \pm 2$ & $0 \pm 0$ & $16 \pm 3$ & $29 \pm 9$ & $14 \pm 8$ & $19 \pm 1$ & $\textbf{69} \pm 11$ \\
\midrule
\multicolumn{8}{@{}l}{\textit{puzzle-4x4-play-singletask}} \\*
task1-v0 & $32 \pm 9$ & $49 \pm 9$ & $34 \pm 8$ & $54 \pm 9$ & $50 \pm 8$ & $72 \pm 8$ & $\textbf{83} \pm 4$ \\*
task2-v0 & $5 \pm 3$ & $4 \pm 4$ & $16 \pm 5$ & $\textbf{17} \pm 5$ & $9 \pm 2$ & $4 \pm 2$ & $5 \pm 0$ \\*
task3-v0 & $20 \pm 10$ & $50 \pm 14$ & $18 \pm 5$ & $47 \pm 8$ & $46 \pm 3$ & $47 \pm 10$ & $\textbf{61} \pm 7$ \\*
task4-v0 $\star$ & $5 \pm 1$ & $21 \pm 11$ & $11 \pm 3$ & $19 \pm 5$ & $17 \pm 3$ & $10 \pm 3$ & $6 \pm 2$ \\*
task5-v0 & $4 \pm 3$ & $2 \pm 2$ & $7 \pm 3$ & $11 \pm 5$ & $7 \pm 3$ & $2 \pm 1$ & $3 \pm 1$ \\
\midrule
\multicolumn{8}{@{}l}{\textit{antmaze-$\ast$-v2 (D4RL)}} \\*
umaze & $94 \pm 2$ & $92 \pm 6$ & $96 \pm 2$ & $98 \pm 0$ & $96 \pm 1$ & $96 \pm 2$ & $\textbf{100} \pm 1$ \\*
umaze-div & $82 \pm 6$ & $62 \pm 12$ & $89 \pm 5$ & $67 \pm 0$ & $91 \pm 2$ & $86 \pm 9$ & $\textbf{96} \pm 5$ \\*
medium-play & $77 \pm 7$ & $56 \pm 15$ & $78 \pm 7$ & $76 \pm 2$ & $81 \pm 5$ & $81 \pm 5$ & $89 \pm 5$ \\*
medium-div & $77 \pm 6$ & $60 \pm 25$ & $71 \pm 13$ & $66 \pm 1$ & $61 \pm 20$ & $75 \pm 8$ & $84 \pm 10$ \\*
large-play & $32 \pm 21$ & $55 \pm 9$ & $84 \pm 7$ & $17 \pm 4$ & $82 \pm 5$ & $83 \pm 6$ & $\textbf{89} \pm 8$ \\*
large-div & $20 \pm 17$ & $64 \pm 8$ & $83 \pm 4$ & $26 \pm 7$ & $84 \pm 5$ & $84 \pm 6$ & $\textbf{85} \pm 5$ \\
\end{longtable}
\endgroup

\end{document}